\documentclass[conference]{IEEEtran}
\IEEEoverridecommandlockouts 

\usepackage[utf8]{inputenc}
\usepackage[T1]{fontenc}
\usepackage{lmodern}
\usepackage{textcomp}
\usepackage{microtype}

\usepackage{amsmath, amssymb, amsfonts}
\usepackage{bm} 
\usepackage{siunitx}
\DeclareSIUnit{\COtwoe}{\ensuremath{\mathrm{CO_2e}}}
\DeclareSIUnit{\gCOtwoe}{\ensuremath{\text{g\,CO_{2e}}}}
\DeclareSIUnit{\cent}{\text{\textcent}}

\usepackage{booktabs}    
\usepackage{tabularx}    
\usepackage{array}       
\usepackage[table]{xcolor}
\usepackage{longtable}
\usepackage{pdflscape}    
\usepackage{tabularray}
\usepackage{ragged2e}

\usepackage{listings}
\usepackage{xcolor}
\lstdefinestyle{pystyle}{
  language      = Python,
  basicstyle    = \ttfamily\small,
  keywordstyle  = \color{blue!70!black}\bfseries,
  commentstyle  = \color{green!50!black},
  stringstyle   = \color{orange!70!black},
  numbers       = left,
  numberstyle   = \tiny,
  numbersep     = 6pt,
  frame         = single,
  rulecolor     = \color{black!40},
  tabsize       = 4,
  showstringspaces = false,
  captionpos    = b
}

\usepackage{enumitem}
\newlist{goal}{enumerate}{1}
\setlist[goal,1]{label=\textbf{G\arabic*},leftmargin=1.8em,align=left}

\usepackage{graphicx}
\usepackage{tikz}
\usetikzlibrary{positioning,arrows.meta,shadows,trees,mindmap}
\usepackage{forest}
\usepackage{smartdiagram}
\usepackage[framemethod=tikz]{mdframed}

\forestset{
  skan tree/.style={
    for tree={
      drop shadow,
      text width=3cm,
      grow'=0,
      rounded corners,
      draw,
      top color=white,
      bottom color=blue!20,
      edge={Latex-},
      child anchor=parent,
      anchor=parent,
      tier/.wrap pgfmath arg={tier ##1}{level()},
      s sep+=2.5pt,
      l sep+=2.5pt,
      edge path'={
        (.child anchor) -- ++(-10pt,0) -- (!u.parent anchor)
      },
      node options={ align=center },
    },
    before typesetting nodes={
      for tree={
        content/.wrap value={\strut ##1},
      },
    },
  },
}

\usepackage{hyperref}
\usepackage{url}

\usepackage{comment}
\usepackage{stfloats}     
\usepackage{adjustbox}    
\usepackage{todonotes}    

\usepackage{algorithm}
\usepackage{algpseudocode}

\DeclareUnicodeCharacter{2082}{\textsubscript{2}}

\begin{document}

\bstctlcite{IEEEexample:BSTcontrol}

\title{An MLCommons Scientific Benchmarks Ontology\\
\thanks{FERMILAB-PUB-25-0701-CSAID}
}

\author{
\IEEEauthorblockN{1\textsuperscript{st} Ben Hawks}
\IEEEauthorblockA{\textit{CS and AI Directorate} \\
\textit{Fermilab}\\
Batavia, IL, USA \\
bhawks@fnal.gov}
\and
\IEEEauthorblockN{2\textsuperscript{nd} Gregor von Laszewski}
\IEEEauthorblockA{\textit{Biocomplexity Institute} \\
\textit{University of Virginia}\\
Charlottesville, VA, USA \\
laszewski@gmail.com}
\and
\IEEEauthorblockN{3\textsuperscript{rd} Matthew D. Sinclair}
\IEEEauthorblockA{\textit{CS Dept} \\
\textit{University of Wisconsin-Madison}\\
Madison, WI, USA \\
sinclair@cs.wisc.edu}
\and
\IEEEauthorblockN{4\textsuperscript{th} Marco Colombo}
\IEEEauthorblockA{\textit{Discovery Partners Institute} \\
\textit{University of Illinois Urbana-Champaign}\\
Chicago, IL, USA \\
mcolom4@illinois.edu}
\and
\IEEEauthorblockN{5\textsuperscript{th} Shivaram Venkataraman}
\IEEEauthorblockA{\textit{CS Dept} \\
\textit{University of Wisconsin-Madison}\\
Madison, WI, USA \\
shivaram@cs.wisc.edu}
\and
\IEEEauthorblockN{6\textsuperscript{th} Rutwik Jain}
\IEEEauthorblockA{\textit{CS Dept} \\
\textit{University of Wisconsin-Madison}\\
Madison, WI, USA \\
rnjain@wisc.edu}
\and
\IEEEauthorblockN{7\textsuperscript{th} Yiwei Jiang}
\IEEEauthorblockA{\textit{CS Dept}\\
\textit{University of Wisconsin-Madison}\\
Madison, WI, USA \\
jiang357@wisc.edu}
\and
\IEEEauthorblockN{8\textsuperscript{th} Nhan Tran}
\IEEEauthorblockA{\textit{CS and AI Directorate} \\
\textit{Fermilab}\\
Batavia, IL, USA \\
ntran@fnal.gov}
\and
\IEEEauthorblockN{9\textsuperscript{th} Geoffrey Fox}
\IEEEauthorblockA{\textit{Biocomplexity Institute and CS Dept} \\
\textit{University of Virginia}\\
Charlottesville, VA, USA\\
vxj6mb@virginia.edu}
}

\maketitle

\begin{abstract}

Scientific machine learning research spans diverse domains and data modalities, yet existing benchmark efforts remain siloed and lack standardization.
This makes novel and transformative applications of machine learning to critical scientific use-cases more fragmented and less clear in pathways to impact.  
This paper introduces an ontology for scientific benchmarking developed through a unified, community-driven effort that extends the MLCommons ecosystem to cover physics, chemistry, materials science, biology, climate science, and more.
Building on prior initiatives such as XAI‑BENCH, FastML Science Benchmarks, PDEBench, and the SciMLBench framework, our effort consolidates a large set of disparate benchmarks and frameworks into a single taxonomy of scientific, application, and system-level benchmarks.
New benchmarks can be added through an open submission workflow coordinated by the MLCommons Science Working Group and evaluated against a six-category rating rubric that promotes and identifies high-quality benchmarks, enabling stakeholders to select benchmarks that meet their specific needs.
The architecture is extensible, supporting future scientific and AI/ML motifs, and we discuss methods for identifying emerging computing patterns for unique scientific workloads.
The MLCommons Science Benchmarks Ontology provides a standardized, scalable foundation for reproducible, cross-domain benchmarking in scientific machine learning.  A companion webpage for this work has also been developed as the effort evolves: \href{https://mlcommons-science.github.io/benchmark/}{https://mlcommons-science.github.io/benchmark/}

\end{abstract}

\begin{IEEEkeywords}
benchmark, mlcommons
\end{IEEEkeywords}

\tableofcontents

\clearpage

\section{Introduction}
\label{sec:intro}

\begin{figure*}[b!]
    \centering
    \includegraphics[width=0.95\linewidth]{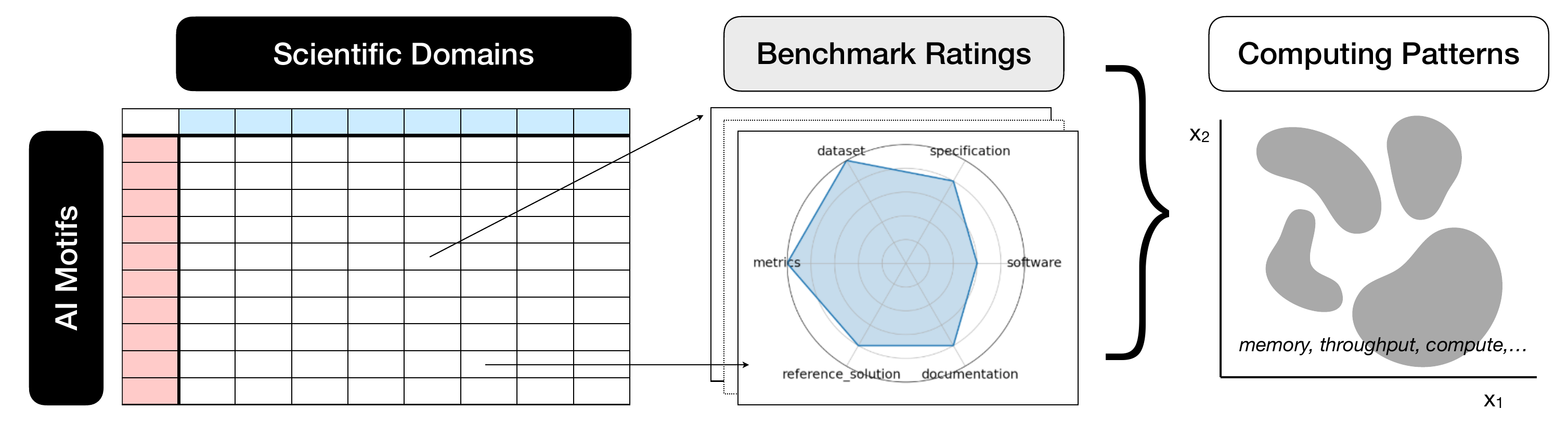}
    \caption{Overview of the Scientific ML Benchmark ontology from the taxonomization of benchmarks by domain and ML motif to the qualification by benchmark through a standardized rating system to the use of the ontology to understand computing patterns for scientific workflows.}
    \label{fig:overview}
\end{figure*}

Benchmarking in scientific machine learning (ML) has emerged as a critical area to guide algorithm development, enable fair comparisons towards progress and innovation, and facilitate reproducibility. The development of ML benchmarks for science is especially critical because of the multi-disciplinary nature of the development, often including domain experts, computing hardware developers, and ML researchers.  That, coupled with the variety of tasks and workloads, makes {\it high quality} benchmarking critical to making progress.  

Our contribution, the MLCommons Science Benchmark Ontology, builds on prior scientific ML benchmarking efforts by integrating their strengths into a unified, community-driven paradigm. Unlike prior domain-specific benchmarking efforts, it provides a unifying principle through an ontology of curated benchmarks across multiple scientific domains. By situating this work under the MLCommons governance model, we ensure long-term community support, extensibility, and adoption, thereby moving from individual benchmark initiatives toward a standardized, widely applicable benchmark ontology for scientific machine learning.


Although each of these benchmarks can be valuable to their communities, there is no standard definition of a high-quality benchmark. This makes accessibility to the problem for different researchers with complementary expertise challenging.  While there is the appearance that there are many ``benchmarks'' for scientific applications in the literature~\cite{CheBeckmann2013, Rodinia, coral2, MattsonCheng2019-mlperfTrain, olcf6-bmks, nersc10, Reddi2020mlperf-Infer, spec, WangPan2017-gunrock}, our analysis finds that their quality varies significantly.  Often, a dataset is provided without concrete tasks, the code is not reproducible, or the metrics are undefined.

Because of this, the state of scientific benchmarking presents a challenge: the vast number and diversity of scientific workloads makes finding a well-defined, high-quality benchmark that targets a given domain a time consuming task. And for some stakeholders who are targeting a broader range of benchmarks, it is infeasible to run all of these benchmarks simply due to the sheer volume. 

A given stakeholder may also have different priorities in searching for benchmarks. For example, hardware vendors typically utilize benchmarks to understand the demands of state-of-the-art applications on potential future systems and ensure these systems provide high performance for important workloads. Researchers and domain scientists use these benchmarks to study how potential hardware and software optimizations will apply to current and/or future systems, and how different solutions to common domain-specific problems compare against others in a standardized format.
Nevertheless, since different benchmarks from different domains value and stress different things, finding a way to identify relevant, representative subsets for different stakeholders is imperative.

Our goal is to develop a standard definition of benchmarking that can be used by the scientific community for ML benchmarking.  They will be classified by different domains and ML motifs.  We will design an accompanying web portal that is searchable based on benchmark quality, domain, and ML motif.  Then we demonstrate an example of how these benchmark tasks can be used to understand emerging computational patterns for scientific HPC.  This overall idea is illustrated in Fig.~\ref{fig:overview}.


\subsection{Related Work}
\label{subsec:intro-relWork}

As shown in \autoref{tab:related-work}, prior efforts have addressed aspects of this challenge.
However, they have often done so within narrow domains or specialized contexts.
Explainability-focused benchmarks such as XAI-BENCH~\cite{xai-bench-2021} provide synthetic datasets with known ground-truth feature importance to systematically evaluate explainable AI methods. While highly valuable for interpretability studies, the scope of XAI-BENCH is restricted to explainability metrics rather than end to end scientific ML workloads.

Similarly, FastML Science Benchmarks~\cite{duarte2022fastml} address an orthogonal need, namely ultra low latency ML in scientific applications such as particle physics event tagging and accelerator control. These benchmarks emphasize hardware/software codesign and real time inference, but they are domain specific and do not generalize across the wide range of scientific machine learning tasks.

A more domain specialized but broader dataset contribution is PDEBENCH (Takamoto et al., 2022)~\cite{takamoto2022pdebench}, which provides large-scale benchmarks for surrogate modeling of partial differential equations (PDEs). Covering forward and inverse problems with diverse PDEs, PDEBENCH offers extensible datasets and APIs along with physics aware evaluation metrics. However, its focus remains limited to PDE based simulations, without addressing broader scientific domains or end to end benchmarking pipelines.

In contrast, SciMLBench~\cite{ThiyagalingamShankar2022-mlSci} introduces a conceptual framework for scientific ML benchmarking. It distinguishes between scientific, application, and system-level benchmarks, and emphasizes the importance of curated datasets, reproducibility, and community standards. While influential in defining the benchmarking landscape, SciMLBench functions more as a blueprint than an implemented suite of benchmarks.

\begin{table*}

\centering
\caption{Comparison of prior related works to this work}
\begin{tblr}{
  width = \linewidth,
  colspec = {Q[138]Q[88]Q[119]Q[231]Q[358]},
  hlines,
  vlines,
  hline{1,7} = {-}{0.08em},
}
Work & Focus Area & Scope & Dataset/Tasks & Key Contributions\\
XAI-Bench~\cite{xai-bench-2021} & Explainable AI benchmarking & Feature attribution & Synthetic datasets with ground-truth feature importance & Provides controllable synthetic datasets for evaluating explainers; narrow focus on interpretability rather than general scientific ML\\
FastML Science Benchmarks~\cite{duarte2022fastmlsciencebenchmarksaccelerating2} & Edge ML for physics & Real time, low latency ML workloads & Tasks in HEP jet tagging, sensor data compression, accelerator control & First to emphasize hardware-software co-design and real-time constraints in scientific ML; scope limited to latency-critical workloads\\
PDEBench~\cite{takamoto2022pdebench} & PDE-based surrogate modeling & Scientific ML for PDE emulation & 11 PDEs (1D–3D: Burgers, Navier–Stokes, shallow-water, Darcy flow, etc.) & Large-scale PDE datasets for forward + inverse problems; extensible APIs; baselines (FNO, U-Net, PINNs)\\
SciMLBench ~\cite{ThiyagalingamShankar2022-mlSci} & General benchmarking framework & Cross-domain ML benchmarks & Datasets from Astronomy, particle physics, materials, and life sciences & Proposes taxonomy: scientific, application, and system benchmarking; emphasizes community standards, curated datasets\\
\textbf{This Work} & End to end scientific ML benchmarking & Broad, community-driven suite integrating diverse workloads & Covers multiple scientific domains (e.g., physics, chemistry, materials, life sciences) with curated benchmarks emphasizing quality & Extends prior work by unifying domain-specific efforts into a standardized, extensible benchmark ontology under MLCommons; stresses reproducibility, FAIR data, community governance, and broad accessibility beyond single-domain focus
\label{tab:related-work}
\end{tblr}
\end{table*}



\section{Scientific Benchmark Definition}
\label{sec:benchmark-definition}

A \textbf{benchmark} is a carefully defined \textit{standardized} version of a scientific application that is used for making quantifiable comparisons of solutions. While many elements below are commonly identified in the literature, they provide the base on which we can structure our work. 

\subsection{Problem Specification and Constraints}
\label{subsec:benchmark-definition-spec}


A succinct statement of the benchmark \textit{task} describes the type of input data, the expected output, and any constraints on the task.  An example of the input description could include its origin or representation, such as image data, time-series data, 3D point cloud, or natural language.  The expected output describes the transformation of the data, such as a regression task, anomaly detection, or generative model.  

The system \textit{constraints} are quantifiable elements of the benchmark (power, latency, etc.) or system specifications (hardware platform, technology node, etc.) that are not a part of the performance metric(s) being compared.  System constraints are typically bounds with an upper or lower limit.  


\subsection{Dataset}
\label{subsec:benchmark-definition-data}

The dataset is the scaffolding which includes input data and potentially truth labels if a part of the benchmark. It adheres to the FAIR principles: each instance is uniquely identified and documented for \textit{findability}; \textit{access} is provided through persistent, open protocols; formats and metadata follow community standards to ensure \textit{interoperability}; and versioning with associated preprocessing scripts enables full \textit{reproducibility} of results. The dataset is bounded: no data augmentation, enrichment, or post hoc curation is permitted unless otherwise specified, ensuring a stable target for comparative evaluation. A canonical, non-overlapping split into training, validation, and test sets is provided to facilitate standardized model development and unbiased performance reporting.


\subsection{Performance Metric(s)}
\label{subsec:benchmark-definition-perf}

The performance metrics are quantifiable measures for comparison.  There is an important distinction between constraints (fixed bounds) and performance metrics (measurable bases for comparison).  A benchmark can have multiple \textit{dimensions} such that more than one metric can be optimized simultaneously that forms a Pareto optimization.  

For example, measures such as mean squared error (MSE) over a held-out test set, relative error in conserved quantities, computational cost (CPU/GPU time), or memory footprint can all be considered.  Sometimes benchmarks also track stability or robustness (e.g. performance when noise is added).  The metrics must be well-defined, reproducible, and measurable when constructed from the validation dataset.  

Multiple \textbf{benchmarks} may be defined for a given \textbf{task} based on its performance metrics.  For example, for the \textit{same} dataset, one benchmark may require a lower bound on accuracy constraint while testing hardware throughput and another benchmark may only test for the best algorithm accuracy.  


\subsection{Reference Solution}
\label{subsec:benchmark-definition-ref}

A reference solution satisfies the problem specification and constraints using the defined dataset and includes performance metrics measurements as a baseline to which other solutions can be compared. 

\subsection{Documentation and Reproducible Protocol}
\label{subsec:benchmark-definition-doc}

A benchmark must include a protocol to reproduce the reference solution.  This will enable additional solutions to make direct comparisons.  Additional proposed solutions must also be reproducible.  
This is typically through reference code within a well-defined software environment that is versioned.  Hardware references must include thorough documentation and a bill of materials.

\section{Existing Benchmarks, Rating and Endorsement System}
\label{sec:currBmks}

In curating the benchmark ontology, we identify numerous existing scientific benchmarks and attempt to categorize and rate them in order to compile them into a comprehensive benchmark ontology.
Our goal in curating this ontology is not to create a ontology that is intended to be exhaustively and completely run, but instead to create a collaborative, high quality benchmark ontology that can be referenced and sampled from when a potential submitter might be searching for domain specific workloads, varying computing motifs, focus areas, and the next generation of challenging machine learning tasks.

However, modern workloads, especially in domains like ML, are evolving rapidly.
Accordingly, our proposed benchmark ontology cannot remain static if we hope to continue representing the state-of-the-art.
Thus, we allow new benchmarks to be proposed and potentially added to the benchmark ontology through an open submission process, reviewed by the MLCommons Science Working Group, and integrated into the ontology upon meeting evaluation standards.
This model ensures that the benchmark remains both extensible and sustainable.

In order to determine quality benchmarks across different scientific domains, we have developed a domain agnostic rating system that can be used to identify quality benchmarks across six different categories, each aligning with a different aspect from the definition of the a benchmark as given in Section~\ref{sec:benchmark-definition}. 

Each category is given a score out of a possible 5 points.
A benchmark may be scored a 0 in a category if none of the criteria for a given category are met, and a 5 if all the criteria for a given category are met.
If a given benchmark obtains an average score across all 6 categories of at least 4.5 out of 5, it is given the ''MLCommons Science Benchmark Endorsement", identifying it as a particularly high quality benchmark within the wider ontology. 

The six categories that are used to evaluate a given benchmark, and their evaluation criteria, are as follows:

\subsection{Software Environment}
\label{subsec:currBmks-sw}

When evaluating the quality of a given benchmark's software environment, we award it a score based on the following criteria, giving it one point for each statement that is true:

\begin{itemize}
   \item The code is available to reproduce the baseline\\reference solution. 
   \item The provided code is complete.
   \item The code itself is well documented
   \item The code does not require any modifications to run.
   \item The environment is either containerized, or environment details and setup instructions are provided.
\end{itemize}

\subsection{Problem Specification and Constraints}
\label{subsec:currBmks-spec}

When evaluating the quality of a given benchmark's problem specification and system constraints, we award it points based on the following criteria, giving it one point for each statement that is true:

\begin{itemize}
    \item System constraints, such as required power, latency, and/or throughput requirements, are provided. 
    \item The benchmark task is clear.
    \item The dataset format is clearly specified. 
    \item The task inputs are clearly specified. 
    \item The task outputs are clearly specified. 
\end{itemize}

\subsection{Dataset}
\label{subsec:currBmks-data}

We evaluate the quality of a given benchmark's dataset by primarily judging how well it adheres to the FAIR Dataset Principles\cite{wilkinson2016fair}, giving a point for each of the 4 principles that are followed, and an additional point is given for if there are well defined train/test(/validation) splits present. 

\subsection{Performance Metrics}
\label{subsec:currBmks-perf}

We evaluate the quality of a given benchmark's performance metrics by judging how well defined the metrics are, and how well they capture a given solutions performance. Points are awarded on the following scales, combining each scale's rating into a single rating for performance metrics overall:

Metric Definitions:
\begin{itemize}
   \item 3 points if the metrics are fully defined
   \item 2 points if the metrics are clearly mentioned, but specific details/implementations (where applicable) are not defined. 
   \item 1 point if some metrics are mentioned, but not clearly defined as what is being tracked for a given benchmark
   \item 0 points if no metrics are mentioned or defined. 
\end{itemize}

Metric Quality:
\begin{itemize}
   \item 2 Points if the metrics fully capture a given solution's performance
   \item 1 point if the metrics partially capture a given solution's performance
   \item 0 points if the metrics do not capture a given solution's performance
\end{itemize}

\subsection{Reference Solution}
\label{subsec:currBmks-ref}

When evaluating the quality of a given benchmark's reference solution, we award it points based on the following criteria, giving it one point for each statement that is true:

\begin{itemize}
    \item A reference solution is publicly available
    \item The provided reference solution is well documented
    \item All hardware and/or software requirements to run the reference solution are listed
    \item All metrics defined as part of the benchmark are evaluated as part of the reference solution
    \item The baseline solution/model is openly available for study (and if a neural network, the architecture, hyperparameters, and code to train it is provided)
\end{itemize}

\subsection{Documentation}
\label{subsec:currBmks-doc}

When evaluating the quality of a given benchmark's documentation, we award it points based on the following criteria, giving it one point for each statement that is true:

\begin{itemize}
    \item The task is explained and well documented.
    \item The task and benchmark background (scientific and otherwise) is clearly explained.
    \item The motivation for the benchmark is clearly explained.
    \item The evaluation criteria are clearly explained. 
    \item An academic paper about the benchmark exists. 
\end{itemize}

\subsection{Existing Benchmark Ontology}
\label{subsec:currBmks-collection}
We include a full list of benchmarks that are currently included in the ontology in \autoref{tab:benchmarks-list} as a part of the Appendix, including the average score across the 6 evaluation categories. For full details on each benchmark, including individual scores for each category, please see von Laszewski, et al~\cite{benchmark-collection}. 

\section{Motifs}
\label{sec:motifs}

\begin{figure*}
    \centering
    \includegraphics[width=1\linewidth]{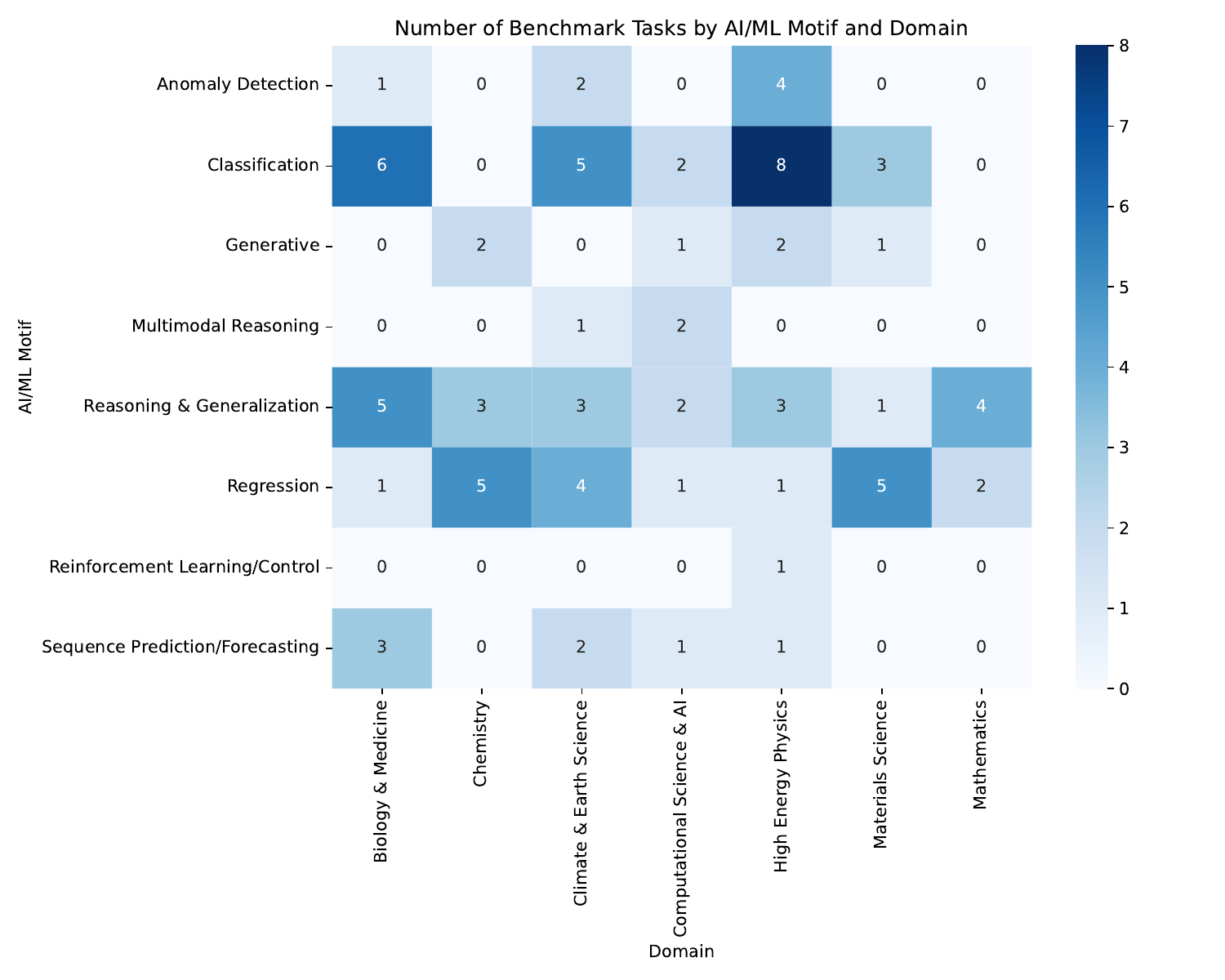}
    \caption{A Heatmap showing the domain and AI/ML Motif for the tasks within the ontology. Note that each task can have multiple domains associated with it, but only a single AI/ML Motif}
    \label{fig:domain-motif-heatmap}
\end{figure*}

\subsection{Primary Users of the Benchmark Ontology}
\label{subsec:motifs-users}

We envision three primary types of users of the benchmark ontology, each having different needs for organizing and finding benchmarks within the ontology.
To make sure that each user is best supported, we've made a website available along side the static report that allows for searching, filtering, and sorting entries within the benchmark ontology. 

\noindent
\textbf{User 1: Specific Benchmark}:
In the first case, we expect a user that is searching for a specific benchmark that matches some given criteria, such as a specific type of AI/ML task within a given scientific domain.
This user is likely searching for a benchmark to test or evaluate a solution to a specific task.
For this user, we've annotated each benchmark with metadata describing it, and have built in a search feature to the website that allows a user to search the metadata for all benchmarks. 
 
\noindent
\textbf{User 2: Specific Benchmark Category}:
In the second case, we expect a user that is searching for a multiple benchmarks within given categories.
This user is likely looking for a subset of benchmarks that are either entirely contained by a single category, such as all biology benchmarks, or some further narrowing of scope, such as all classification tasks within high energy physics.
For this user, we provide a set of tools on the website to filter and sort the benchmark ontology.
 
\noindent
\textbf{User 3: Similar Compute Patterns}:
In the third case, we expect a user that is looking for multiple benchmarks, but not as defined by the given set of categories.
This user might, for example, be searching for computationally similar workloads across multiple AI/ML tasks and scientific domains.
For this user, we present a novel clustering algorithm as an example that can be used to find like benchmarks within the ontology based on the user's specific needs. 

\subsection{Motifs within the ontology}
\label{subsec:motifs-within}

 Due to the broad range of benchmarks within the ontology, we attempt to organize the benchmarks within the ontology by different motifs. We attempt to identify motifs within 2 categories:

\begin{itemize}
\item Scientific Motifs - The scientific domain(s) that a given benchmark centers around (High energy physics, Chemistry, Biology, etc.) 
\item AI/ML Motifs - The type of AI/ML tasks that a given benchmark is trying to accomplish (Classification, Regression, Reasoning, etc.) 
\end{itemize}

We list the broad motifs present in the current iteration of the benchmark ontology below, but intend to add more as it is expanded over time. The domains and motifs that are present in the ontology can additionally be seen in Fig. \ref{fig:domain-motif-heatmap}, which shows the number of tasks present in each domain/motif combination. Note that in the ontology, it is possible for a benchmark task to have multiple domains assocated with it, but only a single AI/ML Motif. Building an exhaustive list fine grained domains and motifs is not possible as this is a constantly evolving ontology, but serves to broadly categorize each benchmark task within the ontology, as well as provide examples on which we can continue to add in order to create a more widely representitive ontology of benchmarks. 

\subsubsection{Scientific Motifs}
\label{subsubsec:motifs-within-sci}

Within the benchmark ontology, we attempt to collect benchmarks that represent a wide variety of scientific domains. 

\paragraph{High‑Energy Physics}

This domain includes benchmarks encompassing particle physics across all aspects of experiment and theory. Particular unique community efforts in~\cite{duarte2022fastml} are identified because they push the limits of inference latency, precision, and hardware‑efficiency on custom accelerator back‑ends.  
Jet‑classification~\cite{duarte2022fastml} and ultrafast jet‑tagging~\cite{odagiu2024ultrafastjetclassificationfpgas} benchmarks demand sub‑microsecond inference on FPGA and quantify both accuracy and resource utilization.
Beam‑control is a reinforcement‑learning problem formulated on a simulated accelerator that tests sample‑efficient policy optimization and stability metrics~\cite{duarte2022fastmlsciencebenchmarksaccelerating3}.
Smart‑pixels~\cite{parpillon2024smartpixelsinpixelai} and Neural Architecture Codesign~\cite{weitz2025neuralarchitecturecodesignfast} further showcase on‑chip inference and low‑latency trigger generation for collider experiments.

\paragraph{Chemistry}

Chemistry benchmarks span generative chemistry, catalytic modeling, and quantum‑chemical corrections.
MOLGEN asks a language model to generate chemically valid molecules in SELFIES and optimize log‑P and docking scores, providing a stringent distribution‑learning test~\cite{fang2024domainagnosticmoleculargenerationchemical}.
OCP (Open Catalyst Project) supplies a vast, DFT‑derived adsorption‑energy dataset that forces regression and graph‑neural‑network models to capture subtle electronic effects~\cite{chanussot2021oc20}.
Delta‑Squared‑DFT adds a second layer of accuracy by training a correction model on CCSD(T) data, testing how well deep networks can interpolate high‑level quantum chemistry from inexpensive DFT~\cite{khrabrov2024nabla2dftuniversalquantumchemistry}.

\paragraph{Materials Science}

Materials‑science benchmarks focus on high‑throughput DFT databases and design‑space exploration.
Materials Project gives a global repository of $>$200\,k inorganic crystal properties that support supervised learning for band‑gap, formation energy, and elastic constants~\cite{jain2013materials}.
JARVIS‑Leaderboard offers an open, community‑driven leaderboard of design‑method comparisons (CGCNN, ALIGNN, M3GNet)~\cite{jarvis}. 
SuperCon3D is a generative benchmark that asks models to predict high‑$T_c$ superconductivity from 3‑D crystal descriptors, combining regression and sampling~\cite{neurips2024_c4e3b55e}.

\paragraph{Biology \& Medicine}

Benchmarks in biology emphasize e data‑driven discovery, multimodal understanding, and biomedical QA.
BiasBench evaluates autonomous scientific discovery by requiring cell‑type annotation and multiple‑choice QA
on scRNA‑seq data~\cite{luo2025benchmarkingaiscientistsomics}.
SeafloorAI couples sonar imagery with natural‑language questions, testing vision‑language grounding in marine geology~\cite{nguyen2024seafloor}.
Vocal‑Call Locator evaluates sound‑source localization on multi‑channel audio in a lab setting~\cite{neurips2024_c00d37d6}.
MedQA is a real‑world medical board‑exam dataset, measuring diagnostic inference and language grounding in clinical knowledge~\cite{jin2020diseasedoespatienthave}.

\paragraph{Climate \& Earth Sciences}

These benchmarks bring remote‑sensing, time‑series, and event‑detection to the fore.
ClimateLearn supplies ERA‑5 based datasets for 3–5‑day weather prediction, testing physics‑aware deep learning~\cite{nguyen2023climatelearnbenchmarkingmachinelearning}.
HDR ML Anomaly Challenge (Sea‑Level Rise) provides satellite‑derived sea‑level time‑series for anomaly detection~\cite{campolongo2025buildingmachinelearningchallenges3},
while SatImgNet offers a multi‑task satellite‑image classification suite~\cite{roberts2023satin}.
MLCommons Science – Earthquake forecasting is a simulation‑based benchmark that evaluates inference speed and prediction accuracy for seismic events~\cite{10.1007/978-3-031-23220-6_4}.

\paragraph{Computational Science \& AI Benchmarks}

These are testbeds that focus on focus on evaluating general performance and throughput of models across multiple domains.
SciCode is a scientist-curated coding benchmark across 16 scientific subfields, evaluating code synthesis for scientific computing tasks.~\cite{tian2024scicoderesearchcodingbenchmark}.
MLPerf HPC benchmarks scientific‑ML training on large‑scale HPC clusters~\cite{farrell2021mlperfhpcholisticbenchmark}.

\paragraph{Mathematics}

Mathematical benchmarks evaluate symbolic reasoning and long‑horizon problem solving.
FrontierMath pushes models on category theory, number theory, and algebraic geometry~\cite{glazer2024frontiermathbenchmarkevaluatingadvanced}.
AIME~\cite{www-aime} and PRM800k~\cite{lightman2023lets} provide structured, multiple‑choice mathematics problems that evaluate generalization across difficulty levels.

\subsubsection{AI/ML Motifs}
\label{sinsubsec:motifs-within-aiML}

\paragraph{Classification}
The classification benchmarks ask a model to assign a discrete label to a given input.
Jet Classification evaluates a model’s ability to distinguish between different particle‑jet types in high‑energy physics data~\cite{duarte2022fastml}.
Smart Pixels for LHC tests on‑chip inference by classifying pixel clusters in a 28nm CMOS detector~\cite{parpillon2024smartpixelsinpixelai}.
BiasBench evaluates cell‑type annotation from single‑cell RNA‑seq data, a multi‑class classification problem~\cite{luo2025benchmarkingaiscientistsomics}.
SeafloorAI asks a model to segment and classify geological features in sonar imagery, while additionally evaluating natural‑language questions~\cite{nguyen2024seafloor}.
SatImgNet is a multi‑task satellite‑image classification suite covering 27 remote‑sensing datasets~\cite{roberts2023satin}.

\paragraph{Regression}
Regression benchmarks assess a model’s ability to predict continuous values.
FEABench measures the accuracy of finite‑element analysis solvers, reporting solve‑time and error norm~\cite{mudur2025feabenchevaluatinglanguagemodels}.
CFDBench evaluates neural‑operator regression on CFD data, reporting L\textsuperscript{2} error and MAE~\cite{luo2024cfdbenchlargescalebenchmarkmachine}.
OCP (Open Catalyst Project) asks for regression of adsorption energies and forces~\cite{chanussot2021oc20}.
Materials Project provides a large DFT database for regression of band‑gap, formation energy, and elastic constants~\cite{jain2013materials}.
Delta Squared‑DFT trains a correction network to map DFT outputs to CCSD(T) energies, evaluated on mean absolute error~\cite{khrabrov2024nabla2dftuniversalquantumchemistry}.

\paragraph{Sequence Prediction / Forecasting}
Sequence‑prediction tasks require a model to forecast future values of a time series.
ClimateLearn provides standardized datasets and evaluation protocols for machine 
learning models in medium-range weather and climate forecasting using ERA5 reanalysis~\cite{nguyen2023climatelearnbenchmarkingmachinelearning}.
GB-Biology is a suite of large-scale biological network datasets (protein-protein
interaction, drug-target, etc.) with standardized splits and evaluation protocols 
for node, link, and graph property prediction tasks.~\cite{hu2021opengraphbenchmarkdatasets}.

\paragraph{Anomaly Detection}
Anomaly‑detection benchmarks test a model’s ability to flag outliers or rare events.
HDR ML Anomaly Challenge (Sea‑Level Rise) uses satellite‑derived sea‑level time‑series to detect flooding anomalies, evaluated with ROC‑AUC and precision/recall~\cite{campolongo2025buildingmachinelearningchallenges3}.
HDR ML Anomaly Challenge (Gravitational Waves) detects anomalous gravitational‑wave signals from LIGO/Virgo data, measured by ROC‑AUC and precision/recall~\cite{campolongo2025buildingmachinelearningchallenges}.

\paragraph{Reinforcement Learning / Control}
Reinforcement‑learning benchmarks involve learning a policy to control a system.
Beam Control trains an RL agent to stabilize accelerator beam trajectories, evaluated on stability loss and control‑latency~\cite{duarte2022fastmlsciencebenchmarksaccelerating3}.
Intelligent Experiments through Real‑time AI uses FPGA‑based trigger generation for sPHENIX/EIC, measured by accuracy and latency~\cite{kvapil2025intelligentexperimentsrealtimeai}.
Quench Detection employs RL and autoencoders to detect superconducting magnet quenches in real time, evaluated by ROC‑AUC and detection latency~\cite{quench2024}.

\paragraph{Generative}
Generative benchmarks ask a model to produce novel data samples.
MOLGEN generates chemically valid molecules in SELFIES, optimizing log‑P, QED, and docking scores~\cite{fang2024domainagnosticmoleculargenerationchemical}.
SuperCon3D generates high‑Tc superconductors from 3‑D crystal descriptors, evaluated by MAE and validity~\cite{neurips2024_c4e3b55e}.
Delta‑Squared‑DFT generates corrected DFT energies, evaluated by mean absolute error~\cite{khrabrov2024nabla2dftuniversalquantumchemistry}.

\paragraph{Multimodal Reasoning}
Multimodal benchmarks combine vision, language, and sometimes audio.
SPIQA (Scientific Paper Image QA) requires a model to answer questions about scientific figures, evaluated by accuracy and F1~\cite{zhong2024spiqa}.
SeafloorAI pairs sonar imagery with natural‑language questions, measured by segmentation pixel accuracy and QA accuracy~\cite{nguyen2024seafloor}.

\paragraph{Surrogate Modeling}
Surrogate‑modeling benchmarks replace expensive traditional simulations with neural surrogates.
CFDBench trains neural operators to emulate CFD, evaluated with L\textsuperscript{2} error and MAE~\cite{luo2024cfdbenchlargescalebenchmarkmachine}.
The Well provides 16 different datasets across multiple scientific domains to evaluate physical surrogate surrogate models~\cite{neurips2024_4f9a5acd}.

\paragraph{Reasoning \& Generalization}
Reasoning benchmarks probe a model’s ability to perform logical inference and extrapolate.
MedQA presents multiple‑choice medical board‑exam questions, measuring diagnostic accuracy~\cite{jin2020diseasedoespatienthave}.
FrontierMath tests advanced mathematical reasoning across category theory, number theory, and algebraic geometry, evaluated by accuracy~\cite{glazer2024frontiermathbenchmarkevaluatingadvanced}.
AIME evaluates high‑school mathematics problem solving, measured by accuracy~\cite{www-aime}.

\subsection{Emerging Computing Patterns}
\label{subsubsec:motifs-comp}

\subsubsection{Computing Motifs}




Computational motifs across different benchmarks are somewhat more difficult to identify, as they tend to be a property of a given implementation rather than the benchmark itself. Nevertheless, some benchmarks are broadly categorizable based on specific system constraints, task architectures, and/or typical computational patterns that a benchmark exhibits. While we do attempt to do so in this work, the categories we identify can differentiate benchmarks by the typical computational area the task is bound by, such as latency, throughput, utilization, or memory. It is also possible that a benchmark can be bound by multiple of these categories. 

\begin{itemize}
\item \textbf{Latency Bound}
Latency-bound benchmarks are those where there is either a fixed latency constraint, or the objective is simply to produce an output as fast as possible. 

\item \textbf{Memory Bound}
Memory-bound benchmarks typically involve tasks that require working with large amounts of data all at once.

\item \textbf{Throughput Bound}
Throughput-bound benchmarks have the need to be able to consume and produce data at very high rates, sometimes independent of the total system latency, typically due to bandwidth constraints present in a system. 

\item \textbf{Utilization Bound}
Utilization-bound benchmarks are benchmarks that have a fixed limit on the amount of computational power they can use, or where the task is traditionally a computationally expensive task, a classical example of which is be protein folding simulations.
\end{itemize}

\subsubsection{Clustering Algorithm}
\label{subsec:motifs-cluster}

\begin{figure*}
    \centering
    \vspace{1ex}
    \includegraphics[width=1\linewidth]{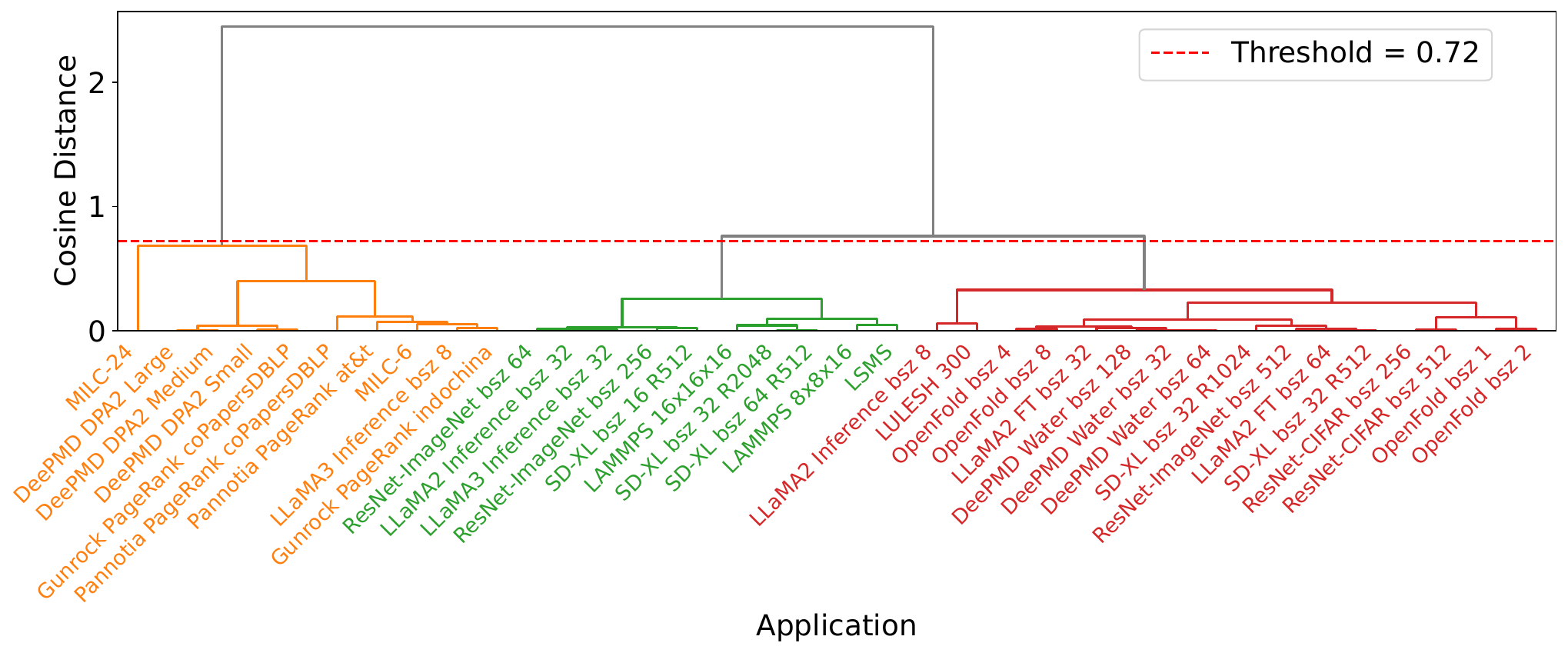}
    \vspace{-3ex}
    \caption{Dendrogram showing hierarchical clustering based on power distributions of workloads. The clusters are labeled Low-power (orange), High-power (green), and Mixed (red), respectively, based on their power distribution.}
    \label{fig:power-cluster}
    \vspace{-1ex}
\end{figure*}


Depending on the primary user's area of expertise and goals when using MLCommons Science benchmarks, they will likely have different priorities in what they view as most important in a set of benchmarks.
Some users will also have additional subgoals beyond a specific benchmark category.
For example, ML practitioners typically prioritize the accuracy (e.g., error) of algorithms.
Similarly, others care about how workloads apply to different compute scenarios (e.g., edge computing) and domain scientists (similar to \textbf{User 2}) prioritize having a variety of representative workloads for their area.
Alternatively, \textbf{User 2} may be interested in multiple different implementations for the same benchmark (\textit{benchmark solutions}).
Conversely, computer systems researchers and hardware vendors (similar to \textbf{User 3} and the \textbf{Computing Patterns}) prioritize ensuring that benchmarks cover a wide variety of architectural or system features (e.g., compute- and memory-intensive workloads, workloads that utilize different architectural features like AMD's Matrix Core Engines~\cite{LohSchulte2023-mi250, SmithLoh2024-mi300A} or NVIDIA's TensorCores~\cite{ChoquetteGiroux2018-volta, ampere, hopper, Choquette2023-hopper}.
Notably, unlike some other users, systems researchers and vendors may be less interested in covering different domains -- unless those domains have applications with different computational patterns they need to optimize their systems for.
More broadly, these needs often can be categorized into motifs (Section~\ref{subsec:motifs-within}).

Naturally, these conflicting goals and motifs make it challenging for a single benchmark ontology to be fully representative for all stakeholders.
Thus, instead of attempting to design a single benchmark ontology that is representative for all stakeholders, we instead propose to design an algorithm that views the priorities of the different stakeholders as \textit{axises} in a multi-dimensional space.
Given input from the user about what their priorities are, our algorithm presents a representative subset of benchmarks or benchmark solutions of interest given those priorities.

Accordingly, we propose to create a workload classifier that clusters the benchmarks, including multiple benchmark solutions for the same application, in our ontology.
By clustering the benchmarks solutions, we can identify workloads with similar behavioral characteristics without requiring expensive system-level profiling for every new application or input per application.
Such a classification scheme can be then used to obtain information about a new workload's behavior, including accuracy (error), compute platform, domain, power consumption, and performance.
To achieve this, we will integrate profiling and clustering to construct a comprehensive view of a given workloads behavior in line with Figure~\ref{fig:overview}.

For each axis (e.g., dataset, specification, software, documentation, reference solution, and performance metrics) we will use the rating system from Section~\ref{sec:currBmks} and user input about which axises are important to them to develop a clustering algorithm that emits the benchmark(s) most relevant to them.
Given this information, we will leverage it to create a \textit{selection bar} for the MLCommons website similar to Figure~\ref{fig:overview}.
Specifically, a user will input how important each of the five axises are.
This information will be passed to our algorithm, which will generate clustering information given these preferences and output the appropriate, representative workloads of interest.
This information will then be displayed on the MLCommons website for a user to use and identify their desired benchmark(s).

The data we need to gather for each axises will vary.
For example, for performance and power metrics, we will leverage hardware profiler metrics (e.g., GPU resource utilization and instantaneous power metrics).
Other metrics like documentation and reference solution will utilize static, offline information.
Once we have gathered this information, we will effectively have an $N$ dimensional vector per workload, where $N$ represents the number of axises.
A smaller $N$ creates coarser grained bins which are easier to group but may group dissimilar applications together if their coarse, aggregate values look similar.
Conversely, a larger $N$ creates finer grained bins, enabling our approach to distinguish fine-grained power level variations, but also more aggressively separating workloads. 

To identify similarities in behavior across workloads represented by these vectors, we apply \textbf{Hierarchical Clustering} to the collected feature vectors.
Hierarchical clustering is an agglomerative classification, starting with leaf nodes where every application is its own cluster, and merging similar pairs of clusters together. 
To do this, the technique groups clusters using a distance metric such as Euclidean or cosine distance. 
We use cosine distance to compute pairwise distance between feature vectors since prior work~\cite{cosinedistance,XIA201539} suggests Euclidean distances are biased towards the magnitude of the feature vectors rather than the direction, and cosine similarity does not suffer from this bias.
Like Fathom~\cite{adolf2016fathom}, we also propose to use hierarchical clustering to group workloads with similar dynamic power signatures.

Figure~\ref{fig:power-cluster} shows an example of how we propose to do our clustering.
Here, we analyze a variety of modern graph analytics, HPC, HPC+ML, and ML GPU workloads~\cite{coral2, Reddi2020mlperf-Infer, WangPan2017-gunrock, mlperf-training, openfold2, olcf6_benchmarks, WangZhang2018-deepmd, CheBeckmann2013-pannotia, touvron2023llama2openfoundation, grattafiori2024llama3herdmodels, EISENBACH2017lsms, thompson2022lammps}, including different inputs for some of these workloads.
For simplicity, here we refer to a given workload with different inputs as a separate benchmark solution.
We profiled each benchmark solution's power consumption using NVIDIA's \texttt{nvidia-smi} and \texttt{NVML} tools.
Given these values, we then calculate the cosine distance between each pair of benchmark solutions.
Figure~\ref{fig:power-cluster}'s dendogram then hierarchically clusters the benchmark implementations based on these values.
The dendrogram's y-axis indicates the cosine distance between two workloads.
A cosine distance of 0 indicates perfectly aligned power consumption while a larger cosine distance indicates that benchmark solutions are farther apart.
Thus, we can slice the dendrogram at suitable cosine distances, to obtain $K$ different groups or clusters.
For example, slicing the dendrogram at a cosine distance of 0.72 yields three distinct power behavior groups ($K$=3).
While this example only uses a single type of information, and different dendrograms can be generated by using different cosine distances, it nonetheless serves as a useful demonstration of how our clustering can be used to produce subsets of benchmark solutions for Figure~\ref{fig:overview}.

\section{Summary}
\label{sec:summ}

This work presents the MLCommons Science Benchmark Ontology, a cohesive, community-driven benchmark ontology that extends the MLCommons ecosystem to scientific machine learning workloads. The ontology expands on existing domain-specific efforts, such as the XAI‑BENCH synthetic explainability datasets, the FastML Science Benchmarks for low-latency physics workloads, the PDEBench surrogate modeling benchmarks, and the SciMLBench conceptual framework, into a single, extensible ontology.   

In this work, we introduce a definition and ontology of scientific machine learning benchmarks, where benchmarks are classified and mapped to their scientific domain and machine learning task type (see \ref{tab:benchmarks-list}). New benchmarks are added through an open submission workflow overseen by the MLCommons Science Working Group. Each submission is evaluated against a six category rating rubric (Software Environment, Problem Specification, Dataset, Performance Metrics, Reference Solution, Documentation) that assigns an overall rating and potential endorsement. The six category scoring framework enables stakeholders, researchers, domain scientists, and hardware vendors to identify representative subsets of benchmarks that align with their specific priorities. The ontology supports adding new scientific domains, AI/ML motifs, and computing motifs as it expands, and we additionally illustrate an example method for automated computing workload classification based on power and utilization characteristics.   
     
Collectively, these elements transform a set of isolated domain-specific benchmark initiatives into a standardized, scalable ontology that promotes reproducibility, community participation, and broad applicability across the scientific machine learning landscape. This work positions MLCommons Scientific Benchmark Ontology as a reference point for future extensions and cross-domain benchmarking efforts.

An up-to-date, full report of all benchmarks in the ontology, as well as a searchable and filterable website, is available at \url{https://mlcommons-science.github.io/benchmark/}.

\section*{Acknowledgments}

This manuscript has been authored by FermiForward Discovery Group, LLC under Contract No. 89243024CSC000002 with the U.S. Department of Energy, Office of Science, Office of High Energy Physics.

This work was supported by DOE ASCR Microelectronics Science Research Center Projects, BIA.  This material is based upon work supported by the U.S. Department of Energy, Office of Science, under contract number DE-AC02-06CH11357.

This work was partially supported by the NSF POSE Phase II Award 2303700.

This work has was supported by the DE-SC0023452: FAIR Surrogate Benchmarks and NSF - 2504401– OAC Core: Data I/O CyberInfrastructure grants. 

The portion of this work done at UW-Madison is supported in part by NSF grant CNS-2312688; Advanced Micro Devices (AMD), Inc. under the AMD AI \& HPC Cluster Program; and by the U.S. Department of Energy, Office of Science, Office of Advanced Scientific Computing Research, under Award Number DE-SC-0026036.

\appendix

\clearpage
\onecolumn
{\footnotesize

\begin{longtable}{|p{0.14\textwidth}|p{0.29\textwidth}|p{0.23\textwidth}|p{0.14\textwidth}|}
\caption{Table of all current benchmarks within the collection}
\\ \hline
\textbf{Citation} & \textbf{Domain} & \textbf{AI/ML Motif} & \textbf{Average Rating}  \\ \hline
\endfirsthead
\hline
\textbf{Citation} & \textbf{Domain} & \textbf{AI/ML Motif} & \textbf{Average Rating}  \\ \hline
\endhead
\hline
\multicolumn{4}{r}{Continued on next page} \\
\endfoot
\hline
\endlastfoot
\cite{nguyen2023climatelearnbenchmarkingmachinelearning} & Climate \& Earth Science & Sequence Prediction/Forecasting & \textbf{5.00} \\ \hline
\cite{nguyen2023climatelearnbenchmarkingmachinelearning} & Climate \& Earth Science & Regression & \textbf{5.00} \\ \hline
\cite{nguyen2023climatelearnbenchmarkingmachinelearning} & Climate \& Earth Science & Regression & \textbf{5.00} \\ \hline
\cite{10.1007/978-3-031-23220-6_4} & Climate \& Earth Science & Classification & \textbf{5.00} \\ \hline
\cite{10.1007/978-3-031-23220-6_4} & Climate \& Earth Science & Sequence Prediction/Forecasting & \textbf{5.00} \\ \hline
\cite{10.1007/978-3-031-23220-6_4} & Biology \& Medicine & Classification & \textbf{5.00} \\ \hline
\cite{10.1007/978-3-031-23220-6_4} & Materials Science & Classification & \textbf{5.00} \\ \hline
\cite{allenai:arc} & Computational Science \& AI & Reasoning \& Generalization & \textbf{4.83} \\ \hline
\cite{fang2024domainagnosticmoleculargenerationchemical} & Chemistry & Generative & \textbf{4.83} \\ \hline
\cite{hu2021opengraphbenchmarkdatasets} & Biology \& Medicine & Sequence Prediction/Forecasting & \textbf{4.83} \\ \hline
\cite{zhang2024empowering} & Climate \& Earth Science & Reasoning \& Generalization & \textbf{4.67} \\ \hline
\cite{tian2024scicoderesearchcodingbenchmark} & Computational Science \& AI & Generative & \textbf{4.50} \\ \hline
\cite{krause2024calochallenge2022communitychallenge} & High Energy Physics & Generative & \textbf{4.50} \\ \hline
\cite{takamoto2024pdebenchextensivebenchmarkscientific} & Computational Science \& AI, Climate \& Earth Science, Mathematics & Regression & \textbf{4.50} \\ \hline
\cite{neurips2024_0db7f135} & Climate \& Earth Science & Regression & \textbf{4.50} \\ \hline
\cite{neurips2024_0db7f135} & Climate \& Earth Science & Classification & \textbf{4.50} \\ \hline
\cite{neurips2024_0db7f135} & Climate \& Earth Science & Classification & \textbf{4.50} \\ \hline
\cite{neurips2024_0db7f135} & Climate \& Earth Science & Anomaly Detection & \textbf{4.50} \\ \hline
\cite{pramanick2025spiqadatasetmultimodalquestion} & Computational Science \& AI & Multimodal Reasoning & 4.42 \\ \hline
\cite{karargyris2023federated} & Biology \& Medicine & Classification & 4.33 \\ \hline
\cite{karargyris2023federated} & Biology \& Medicine & Classification & 4.33 \\ \hline
\cite{karargyris2023federated} & Biology \& Medicine & Classification & 4.33 \\ \hline
\cite{nguyen2024seafloor} & Climate \& Earth Science & Classification & 4.33 \\ \hline
\cite{nguyen2024seafloor} & Climate \& Earth Science & Reasoning \& Generalization & 4.33 \\ \hline
\cite{neurips2024_a8063075} & High Energy Physics & Classification & 4.33 \\ \hline
\cite{neurips2024_a8063075} & High Energy Physics & Classification & 4.33 \\ \hline
\cite{neurips2024_a8063075} & Biology \& Medicine & Classification & 4.33 \\ \hline
\cite{neurips2024_a8063075} & Materials Science & Regression & 4.33 \\ \hline
\cite{chanussot2021oc20,tran2023oc22,doi:10.1021/acscatal.0c04525,tran2023b} & Chemistry, Materials Science & Regression & 4.17 \\ \hline
\cite{duarte2022fastml} & High Energy Physics & Classification & 4.17 \\ \hline
\cite{duarte2022fastmlsciencebenchmarksaccelerating2} & High Energy Physics & Generative & 4.17 \\ \hline
\cite{farrell2021mlperfhpcholisticbenchmark} & High Energy Physics & Regression & 4.17 \\ \hline
\cite{farrell2021mlperfhpcholisticbenchmark} & Climate \& Earth Science & Classification & 4.17 \\ \hline
\cite{farrell2021mlperfhpcholisticbenchmark} & Chemistry & Regression & 4.17 \\ \hline
\cite{farrell2021mlperfhpcholisticbenchmark} & Biology \& Medicine & Sequence Prediction/Forecasting & 4.17 \\ \hline
\cite{campolongo2025buildingmachinelearningchallenges} & High Energy Physics & Anomaly Detection & 4.17 \\ \hline
\cite{neurips2024_c4e3b55e} & Materials Science & Regression & 4.17 \\ \hline
\cite{neurips2024_c4e3b55e} & Materials Science & Generative & 4.17 \\ \hline
\cite{luo2025benchmarkingaiscientistsomics} & Biology \& Medicine & Reasoning \& Generalization & 4.00 \\ \hline
\cite{luo2025benchmarkingaiscientistsomics} & Biology \& Medicine & Classification & 4.00 \\ \hline
\cite{neurips2024_4f9a5acd} & Biology \& Medicine, Computational Science \& AI, High Energy Physics & Sequence Prediction/Forecasting & 4.00 \\ \hline
\cite{hendrycks2021measuring} & Computational Science \& AI & Reasoning \& Generalization & 3.83 \\ \hline
\cite{roberts2023satin} & Climate \& Earth Science & Multimodal Reasoning & 3.83 \\ \hline
\cite{rein2023gpqagraduatelevelgoogleproofqa} & Biology \& Medicine, Chemistry, High Energy Physics & Reasoning \& Generalization & 3.83 \\ \hline
\cite{lightman2023lets} & Mathematics & Reasoning \& Generalization & 3.83 \\ \hline
\cite{mudur2025feabenchevaluatinglanguagemodels} & Mathematics & Reasoning \& Generalization & 3.83 \\ \hline
\cite{weitz2025neuralarchitecturecodesignfast} & High Energy Physics & Classification & 3.83 \\ \hline
\cite{khrabrov2024nabla2dftuniversalquantumchemistry} & Chemistry, Materials Science & Regression & 3.83 \\ \hline
\cite{campolongo2025buildingmachinelearningchallenges3} & Climate \& Earth Science & Anomaly Detection & 3.83 \\ \hline
\cite{neurips2024_c00d37d6} & Biology \& Medicine & Regression & 3.83 \\ \hline
\cite{neurips2024_c6c31413} & Chemistry & Generative & 3.75 \\ \hline
\cite{neurips2024_c6c31413} & Chemistry & Regression & 3.75 \\ \hline
\cite{neurips2024_c6c31413} & Chemistry & Regression & 3.75 \\ \hline
\cite{zhong2024spiqa} & Computational Science \& AI & Multimodal Reasoning & 3.67 \\ \hline
\cite{rein2023gpqagraduatelevelgoogleproofqa2} & Biology \& Medicine, High Energy Physics, Chemistry & Reasoning \& Generalization & 3.67 \\ \hline
\cite{jin2020diseasedoespatienthave} & Biology \& Medicine & Reasoning \& Generalization & 3.50 \\ \hline
\cite{diguglielmo2025endtoendworkflowmachinelearningbased} & Computational Science \& AI & Classification & 3.50 \\ \hline
\cite{luo2024cfdbenchlargescalebenchmarkmachine} & Mathematics & Regression & 3.33 \\ \hline
\cite{cui2025curieevaluatingllmsmultitask} & Materials Science, High Energy Physics, Biology \& Medicine, Chemistry, Climate \& Earth Science & Reasoning \& Generalization & 3.33 \\ \hline
\cite{parpillon2024smartpixelsinpixelai} & High Energy Physics & Classification & 3.33 \\ \hline
\cite{https://doi.org/10.5281/zenodo.5046389} & High Energy Physics & Anomaly Detection & 3.33 \\ \hline
\cite{bowles2024betterclassicalsubtleart} & Computational Science \& AI & Classification & 3.17 \\ \hline
\cite{odagiu2024ultrafastjetclassificationfpgas} & High Energy Physics & Classification & 3.17 \\ \hline
\cite{liu2021braggnnfastxraybragg} & Materials Science & Classification & 3.17 \\ \hline
\cite{qin2023extremely} & Materials Science & Classification & 3.17 \\ \hline
\cite{duarte2022fastmlsciencebenchmarksaccelerating3,kafkes2021boostrdatasetacceleratorcontrol} & High Energy Physics & Reinforcement Learning/Control & 3.00 \\ \hline
\cite{kvapil2025intelligentexperimentsrealtimeai} & High Energy Physics & Classification & 3.00 \\ \hline
\cite{campolongo2025buildingmachinelearningchallenges2} & Biology \& Medicine & Anomaly Detection & 3.00 \\ \hline
\cite{abud2021deep} & High Energy Physics & Anomaly Detection & 2.83 \\ \hline
\cite{glazer2024frontiermathbenchmarkevaluatingadvanced} & Mathematics & Reasoning \& Generalization & 2.50 \\ \hline
\cite{www-aime} & Mathematics & Reasoning \& Generalization & 2.33 \\ \hline
\cite{quench2024} & High Energy Physics & Anomaly Detection & 2.17 \\ \hline
\cite{jain2013materials} & Materials Science & Regression & 1.92 \\ \hline
\cite{wei2024lowlatencyopticalbasedmode} & High Energy Physics & Classification & 1.50
\label{tab:benchmarks-list}
\end{longtable}
}
\clearpage
\twocolumn

\bibliographystyle{IEEEtran} 

\bibliography{%
vonLaszewski-frontiers-citations,%
ref-carpentry,%
ref-sinclair,%
benchmarks}

\end{document}